\pdfoutput=1

\documentclass[11pt]{article}

\usepackage{EMNLP2022}

\usepackage{times}
\usepackage{latexsym}
\usepackage{graphicx}
\usepackage{booktabs}
\usepackage{multirow}
\usepackage{bbding}
\usepackage[fleqn]{amsmath}
\newcommand*\samethanks[1][\value{footnote}]{\footnotemark[#1]}
\usepackage[T1]{fontenc}

\usepackage[utf8]{inputenc}

\usepackage{microtype}

\usepackage{inconsolata}

\title{Find Someone Who: Visual Commonsense Understanding in Human-Centric Grounding}

\author{Haoxuan You\textsuperscript{1}, Rui Sun\textsuperscript{1}\thanks{\indent Equal Contribution}, Zhecan Wang\textsuperscript{1}\samethanks[1], Kai-Wei Chang\textsuperscript{2}, Shih-Fu Chang\textsuperscript{1}
\\
\textsuperscript{1} Columbia University, New York\\
\textsuperscript{2} University of California, Los Angeles \\
  \texttt{\{hy2612, rs4110, zw2627, sc250\}@columbia.edu, kwchang@cs.ucla.edu} 
  }

\begin{document}
\maketitle
\begin{abstract}
From a visual scene containing multiple people, human is able to distinguish each individual given the context descriptions about what happened before, their mental/physical states or intentions, \textit{etc}. Above ability heavily relies on human-centric commonsense knowledge and reasoning.  For example, if asked to identify the ``person who needs healing'' in an image, we need to first know that they usually have injuries or suffering expressions, then find the corresponding visual clues before finally grounding the person. 



We present a new commonsense task, \emph{Human-centric Commonsense Grounding}, that tests the models' ability to ground individuals given the context descriptions about what happened before, and their mental/physical states or intentions. We further create a benchmark, \emph{HumanCog}, a dataset with 130k grounded commonsensical descriptions annotated on 67k images, covering diverse types of commonsense and visual scenes. We set up a context-object-aware method as a strong baseline that outperforms previous pre-trained and non-pretrained models. Further analysis demonstrates that rich visual commonsense and powerful integration of multi-modal commonsense are essential, which sheds light on future works. Data and code will be available at \href{https://github.com/Hxyou/HumanCog}{https://github.com/Hxyou/HumanCog}.

\end{abstract}

\section{Introduction}

\begin{figure}[t!]
        \includegraphics[width=0.29\textheight]{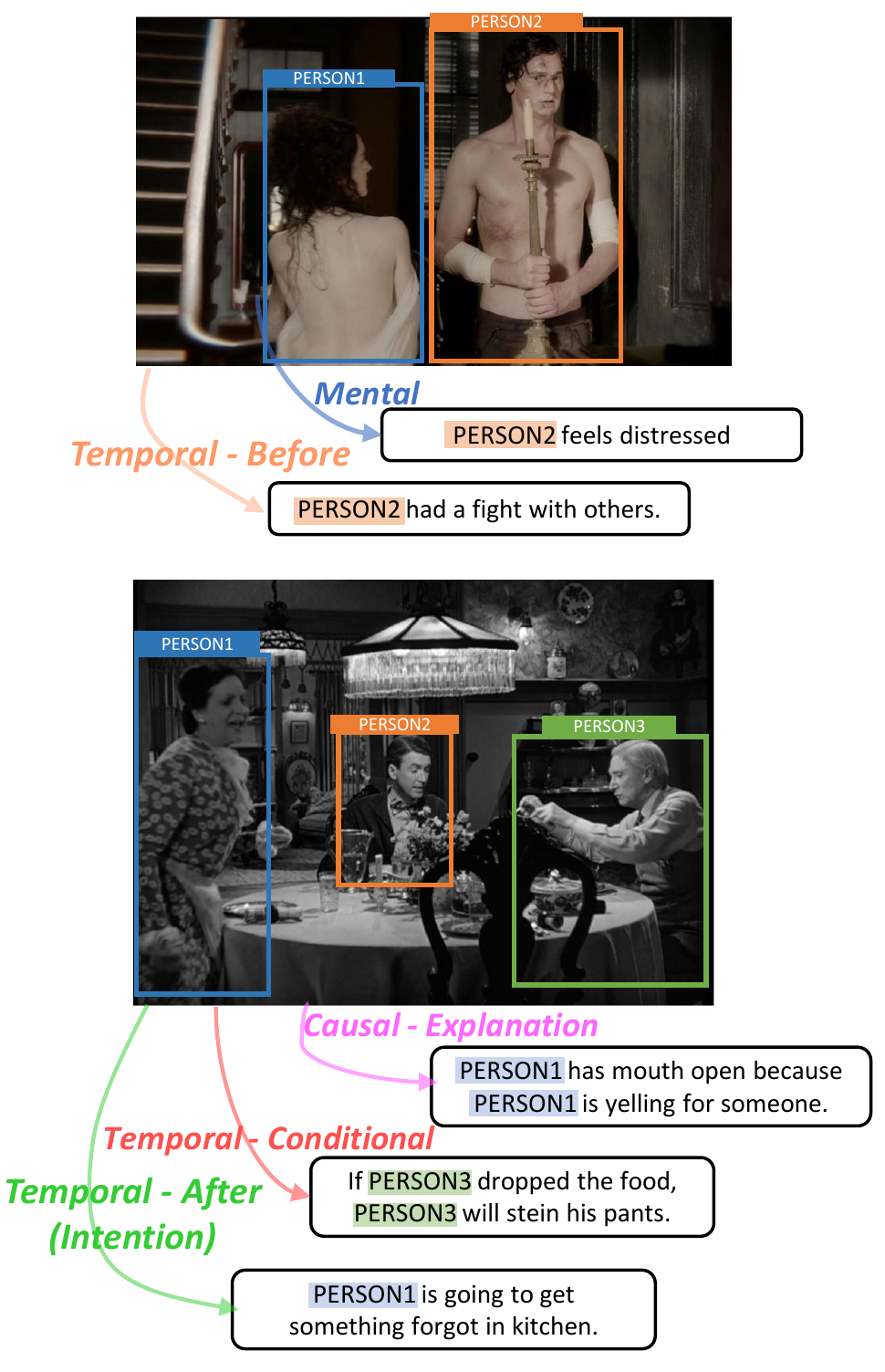}
        \caption{\textit{Human-centric Commonsense Grounding}: given an image, a set of candidate person boxes and a human-centric commonsensical description, a model must ground the persons in description to correct person boxes in image. }
\label{fig:motivation}
\end{figure}

Visual scenes often involve multiple people. For instance, as in movies, a frame can involve multiple characters. Complex human interaction happens because different people may have different intentions, roles, and emotions.
When observing such scenes, humans can understand the scene and differentiate the characters from each other according to the context description, such as what will happen/happened to them, attributes, mental/physical states, and intentions. 

Take the bottom image in Fig. \ref{fig:motivation} as an example. When the context description talks about the figure who is ``going to  get something forgot in the kitchen'', 
humans can match that description to PERSON1 because PERSON1 is standing up, looking around, and speaking. We can also connect dots via the understanding that PERSON2 and PERSON3 are sitting and focusing on eating food without any sign of leaving.
We can achieve this because we understand the commonsense, such as causal, temporal, mental, \textit{etc.}, behind the human interactions and the subtle visual clues can serve as hints to identify them. 



Understanding human-centric commonsense relations is important in widely broad fields. \textit{e.g.} in human-robot communication, it's crucial for medical-aid robots to identify  ``person who needs healing'' and take actions to help. 
Despite the importance and challenges, lack of development exists for this task. Existing works are limited to conventional visual object grounding. The state-of-the-art grounding models can ground objects by the description of their geometric/spatial relation and appearance \cite{mao2016generation, lin2014microsoft, plummer2015flickr30k}, and ground people by actions \cite{cui2021s}, but struggle with complex scenes requiring human-centric commonsense knowledge and sophisticated reasoning ability. Meanwhile, tasks that focus on evaluating commonsense reasoning often take the form of multi-choice QA \cite{zellers2019recognition} or free-form generation \cite{park2020visualcomet}. Such formulation tends to offer less interpretability and could contain easy shortcuts. We formulate our task as a grounding task, with a simple output format (i.e., finding the alignment between humans and bounding boxes) while covering a wide range of commonsense understanding. 


In this paper, we formulate the task as \textbf{Human-centric Commonsense Grounding}. Given an image containing multiple candidate persons and a commonsensical description (including temporal, causal, mental \textit{, etc.} human-interaction), a machine must ground the persons mentioned in the description to correct person boxes in the image. To back the study of this task, we introduce \textbf{HumanCog}, a new dataset with 130k "commonsensical" descriptions where context descriptions are constituted with human-centric commonsense relations like mental states, intentions, \textit{etc.}. Those relations with associated pronouns can all be grounded to 230k persons in 67k images, covering diverse visual scenes. \textit{HumanCog} is automatically collected by transferring the questions and correct answers in Visual Commonsense Reasoning (VCR) to grounded statements through a set of pre-defined rules, which also preserves the paired person-box groundings (\textit{i.e.}, co-reference links). Since the questions in VCR require commonsense to answer, our transferred grounded statements also cover various types of commonsense needed to ground the persons. Further we employ NLP specialists to iteratively refine the rules until reaching a acceptably low error rate. Moreover, the validation and test sets are verified by Amazon Mechanical Turk, and the result testifies the preciseness of our annotation.

We introduce a context-object-aware method as a strong baseline on this task based on pre-trained vision and language Transformer architecture \cite{chen2019uniter}. We take the candidate person region features as weights in classifier and classify the person tokens in text by cross-entropy loss. To facilitate the interaction between people and visual scenes, detected context objects are also input to the model. Further, at the feature level, we draw the person tokens in text and neighbor context objects pertaining to corresponding persons, while push them away from other persons, through a proposed context contrastive loss. 

Comprehensive experiments are conducted with a wide range of methods, from heuristic methods to pre-trained models.  We further present both qualitative and quantitative analysis and find that rich contextualized visual representation, effective usage of context objects, and better integrating vision and text by vision-language pre-training, are the keys to improve the performance. 

In summary, our main contribution is threefold. (1) We introduce a new task, \emph{human-centric commonsense grounding}, to ground the persons mentioned in commonsensical descriptions. (2) A large-scale dataset, \textit{HumanCog}, containing 130k commonsensical descriptions on 67k images. (3) A context-object-aware model to facilitate the visual commonsense learning, establishing a strong baseline on our new challenge. 

\section{Related Work}
\paragraph{Visual Grounding Dataset}
Ground the corresponding regions in images given text information is an essential task to bridge image and text modalities. There are in general two conventional settings in existing grounding datasets. In the first setting, the entire sentence is mainly describing one object and its environment/attribute, and only refers to one box in image, which is termed as Referring expression comprehension (REC). RefCOCO \cite{yu2016modeling}, RefCOCO+ \cite{yu2016modeling} and RefCOCOg \cite{mao2016generation} are commonly used REC datasets annotated on top of MSCOCO \cite{lin2014microsoft}. RefCOCO and RefCOCO+ are annotated using ReferIt Game \cite{KazemzadehOrdonezMattenBergEMNLP14}, but RefCOCO+ focuses more on appearance description since location words are not allowed. RefCOCOg is collected by Mechanical Turks in a non-interactive setting. CLEVR-Ref+ \cite{liu2019clevr} is collected on synthesized images where objects of different attributes are put on plane. KB-REF \cite{wang2020give} enriches the sentence description by injecting knowledge retrieved from the external knowledge base. Flick30k Entity \cite{plummer2015flickr30k} is the pioneer to establish the second setting: multiple phrases inside one caption can be grounded to different boxes. Who's Waldo\cite{cui2021s} further studies grounding the persons in sentence. Our work differs from Who's Waldo and KB-REF in that the descriptions contain rich human-centric commonsense such as temporal, causal, mental, \textit{etc.}  

\paragraph{Visual Grounding Approaches} Current methods can be divided into two categories: one-stage and two-stage\cite{qiao2020referring}. In two-stage methods, a set of candidate object regions are first detected by object detection models, then multimodal models are used to predict the links between detected boxes and text. LSTM-based models \cite{luo2017comprehension, hu2017modeling}, attention-based models\cite{yu2018mattnet, kim2018bilinear, fukui2016multimodal}, graph-based models \cite{liu2020learning, yang2019cross} and pre-trained models \cite{li2019visualbert, lu2019vilbert, chen2019uniter} are explored. In one-stage models, the coordinates of grounded object box are directly predicted by a single model \cite{liao2020real,deng2021transvg, kamath2021mdetr}.More can be found in a survey \cite{qiao2020referring}. Since our dataset already gives ground-truth person candidate boxes, \textit{i.e.}, the first stage results in two-stage schema are provided, we mainly focuses on building better model for image-text understanding (second stage in two-stage schema).

\paragraph{Multimodal Commonsense Reasoning} 
Multimodal commonsense reasoning has attracted wide research interest in recent years. VCR \cite{zellers2019recognition} introduces commonsense question that requires a deep understanding of both image and text, and is formulated as a multi-choice answering task. VisualCOMET \cite{park2020visualcomet} focuses on inferring the temporal and causal information given current image and description, regarded as a generation task. VLEP \cite{lei2020more} also requires machine to predict future event but is in multi-choice answering format. We are similar to VCR in that we collect images from VCR and transform the questions\&answers in VCR to statements. However, we differs from above works in that we target at the human-centric commonsense grounding ability of machines. 

\section{Task: Human-centric Commonsense Grounding}
We present a challenging task, \emph{human-centric commonsense grounding}, to mimic the inference ability of humans to distinguish wanted persons in image by corresponding commonsensical description. The input of one sample in this task includes: (1) An image $\mathnormal{I}$. (2)
 A set of $N$ $(N\geq2)$ candidate person boxes $\boldsymbol{r}$, covering all the  persons in the image. (3) A commonsensical description $\boldsymbol{t}=\{t_i\}_{i=1}^n$ of the image, where $n$ is the token number, \textit{e.g.}, ``PERSONX feels distressed.''  $t_i$ is either a token in vocabulary or a person link (PERSONX in above example) that remains to be grounded/referred to ground-truth person in image.  At least one person link exists.

Given above input, the goal of the task is to ground/refer the person links to corresponding correct person boxes out of all candidate person boxes, \textit{i.e.}, $\mathop{\arg\max}_{\boldsymbol{r}} f(t_i | \boldsymbol{t}, \mathnormal{I}, \boldsymbol{r})=r_j, \{t_i, r_j\} \in \mathcal{L}$, where $\mathcal{L}$ is the set of ground-truth reference pairs and $f$ is the desired model. We evaluate the accuracy of correct prediction among candidate person boxes. 

Take Fig. \ref{fig:motivation} as an example. For the top picture, a commonsensial description is ``PERSONX feels distressed'', where ``PERSONX'' is an person link, and its corresponding ground-truth person in image should be ``PERSON2''. It's noted that there might be more than one person links referring to the same person in image (see the bottom picture in Fig. \ref{fig:motivation}).

\section{Dataset Collection and Analysis}
The \textit{HumanCog} dataset contains 130k commonsensical descriptions on 67k images, where in total 230k persons are grounded. In the following, we describe how \textit{HumanCog} is constructed and annotated, and provide detailed analysis of the dataset. 

\begin{figure*}[t]
        \includegraphics[width=0.65\textheight]{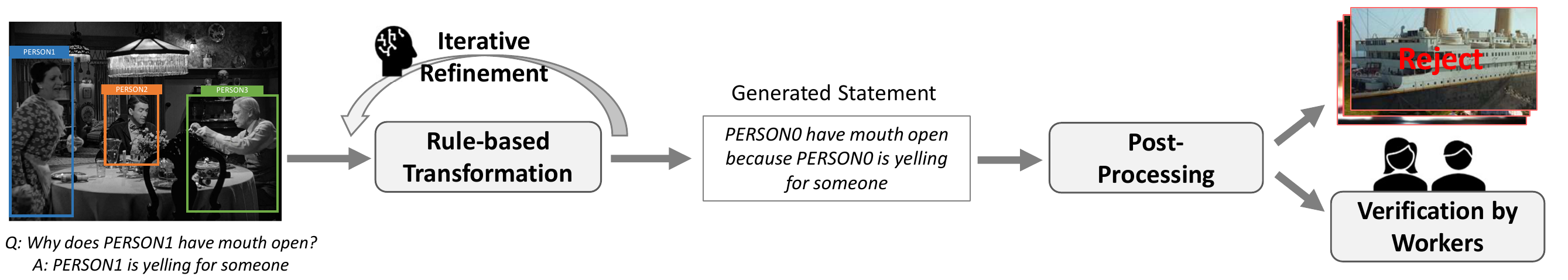}
        \caption{Diagram of data collection.}
\label{fig:annotation}
\end{figure*}

\subsection{Data Collection}
To support the research of human-centric commonsense grounding task, we hope the samples in dataset should have two properties: (1) cover a wide range of visual scenes, (2) have rich and practical commonsense in the description. Although employing annotators to annotate from scratch is a feasible way, it might be costly to build such a large-scale dataset. Instead, we find VCR \cite{zellers2019recognition} a perfect base for our usage, because VCR's images are from movie clips depicting complex and diverse situations, and its questions and answers are carefully annotated by turkers in free-form focusing on various commonsense. To build \textit{HumanCog} dataset, we extend and tailor VCR dataset by following steps: (1) Rule-based Transformation with Iterative Refinement (2) Post-processing (3) Validation with Amazon Turker.

\begin{table*}[t]
\begin{center}
\renewcommand\tabcolsep{3pt} 
\scalebox{0.75}{
\begin{tabular}{cccccccc}
    \toprule

    Dataset &  V. Src. & \#Description & \#Image  & \#Target  & Avg. Word Len. & Human-centric  & Knowledge Reqiured \\
    \midrule
    RefCOCO  & MSCOCO & 142k & 20k & 142k & 3.61 & \XSolidBrush &Spatial/Appearance/Action\\
    RefCOCO+  & MSCOCO & 141k & 20k & 141k & 3.53  &\XSolidBrush &Appearance\\
    RefCOCOg  & MSCOCO & 104k & 27k & 104k & 8.43  & \XSolidBrush &Spatial/Appearance/Action\\
    
    Flickr 30k entities  & Flickr & 159k & 32k & 276k & - & \XSolidBrush &Spatial/Appearance/Action\\
    Who's Waldo  & News & 193k & 193k & 215k & - & \Checkmark & Event/Activity \\
    HumanCog(ours)  & Movies & 130k & 67k & 230k  & 10.32 & \Checkmark &Commonsense-Temporal/Causal/Mental\\
    \bottomrule
\end{tabular}
}
\end{center}
\caption{Comparison with other grounding datasets. `-' denotes data not provided in their paper. Additionally, the average number of people in the image (sentence) in our dataset is 4.11 (1.85).}
\label{tab:datasets}
\end{table*}

\begin{figure}[t]
        \centering
        \includegraphics[width=0.3\textheight]{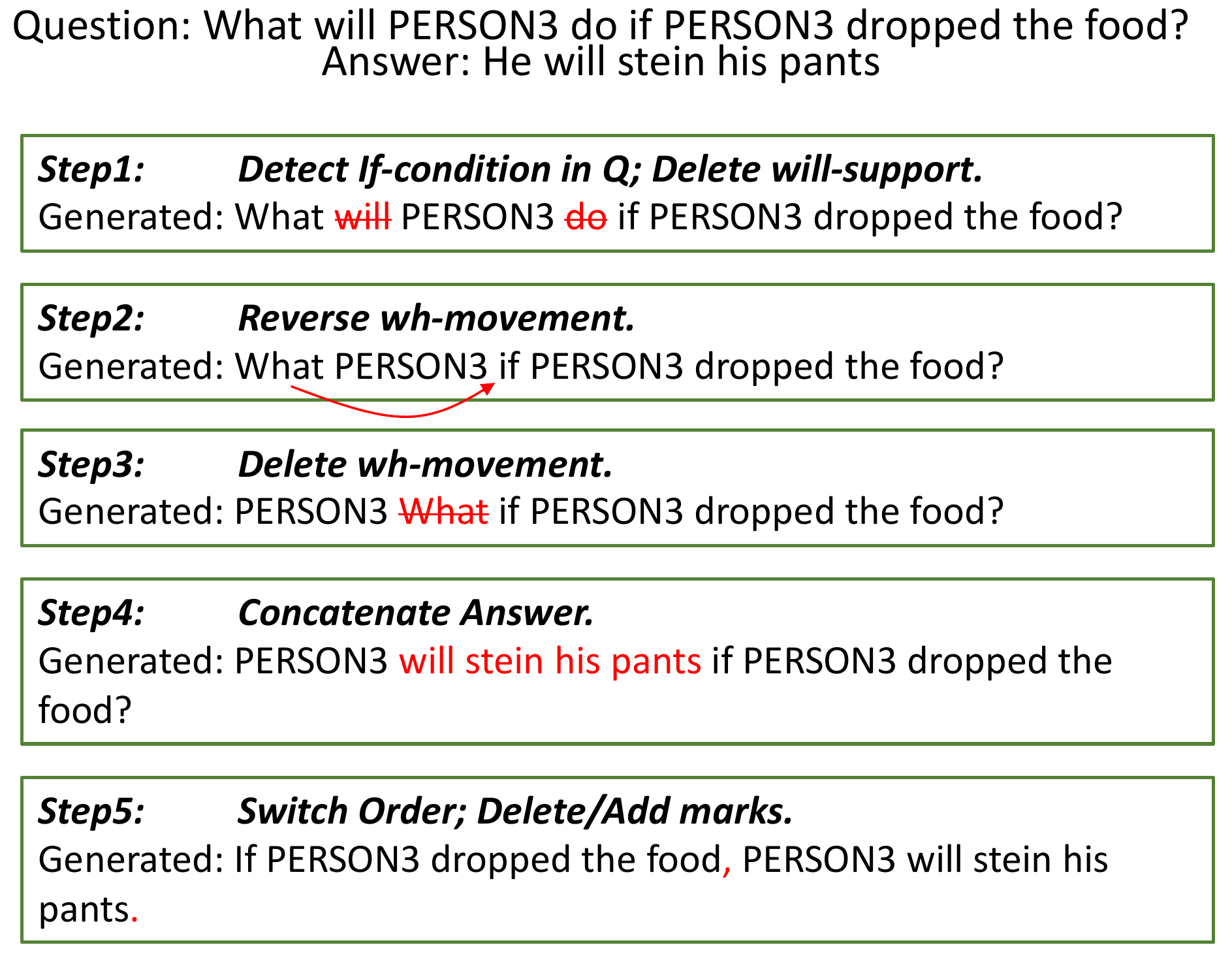}
        \caption{An example of applying one of the rules to a sample.}
\label{fig:template}
\end{figure}

\paragraph{Transformation via rules with Iterative Refinement}
Since the task of VCR is question answering, the text part of each sample in VCR contains one question and four answer choices. To extend VCR for our task, following \cite{demszky2018transforming}, we transform the questions and answers to statements/descriptions via a set of rules. More specifically, in each example, among four answer choices, we take the correct answer $a_{i}^{gt}$ and question $q_{i}$  as input $\{q_{i},  a_{i}^{gt}\}$, since the other answer choices are semantically wrong or irrelevant. Assume we have a set of $Q$ pre-defined rules $\{T_i\}_{i=1}^Q$. During transformation, we first examine whether each sample $\{q_{i},  a_{i}^{gt}\}$ can match certain rule out of all rules. We discard the unmatched examples while keeping the matched ones, which preserves $93.3\%$ samples in train and validation set of VCR (223k out of 239k). Then we transform each $\{q_{i},  a_{i}^{gt}\}$ to a statement via the matched rule. 

 To design different rules, we start from question types and find there are mainly 7 question types in VCR, which begin with \textit{what}, \textit{whose}, \textit{how}, \textit{where}, \textit{who} and \textit{which}. We first define a basic rule for each question type. Then we employ NLP specialists to iteratively refine those basic rules to cover as many as possible various scenarios under each question type. To be more concrete, in every iteration, 20 question-answer pairs per question types are sampled, the NLP specialist have to examine whether current rules can perfectly transform them. If not, they can revise current rules or create new rules for unseen scenarios. Tens of iterations are conducted until current rules can correctly transform all samples in last 5 iterations. In Fig. \ref{fig:template}, We show one example of applying a rule to a question-answer pair in our data. At the end, 15 rules are summarized to do the transformation. The accuracy of our rules are validated by further Amazon Turker Annotation, which will be introduced later. 

\paragraph{Post-processing}
After obtaining the statements, several steps are applied in post-processing. (1) VCR contains all person bounding boxes and object bounding boxes in image, which are already verified by Amazon Turker, and annotated person/object-region co-reference links. Among verified bounding boxes, we only keep the person bounding boxes as candidate person boxes in our task. As for the co-reference links, we keep person-box links as ground-truth grounding labels, and replace the object-region links mentioned in statement with their object names. In that way, each token in statements is either a word in vocabulary or a person link. (2) We remove the samples that have no person links in statements or no person candidate box in images. (3) We remove samples that have only one candidate person box in image, in that the accuracy would be $100\%$ for those samples. (4) Some samples contain too many persons in image, where the persons tend to be blurry and incomplete. To reduce such noise, we remove samples that have more than 10 persons in image. (5) In some cases, there will be two or more person links tied together in description, \textit{e.g.}, ``PERSON1 and PERSON2''. As a result, the person links can be exchanged, which causes the ambiguity. We simply remove those samples.  

In summary, after above post-processing, we keep 134k samples (out of 223k), which are split into 120k training, 7k validation and 7k test set.

\begin{figure}[t]
        \centering
        \includegraphics[width=0.3\textheight]{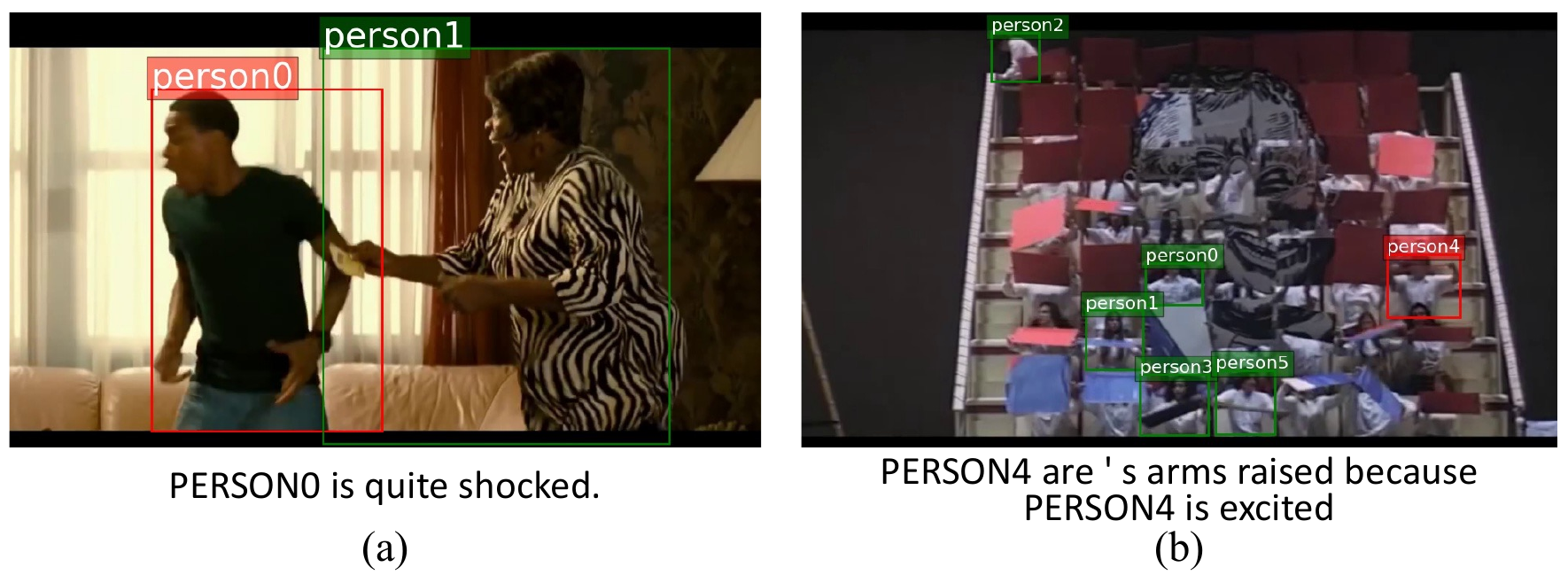}
        \caption{Ambiguous and unreadable examples.}
\label{fig:turker}
\end{figure}

\paragraph{Validation with Amazon Turker}
Workers on Amazon Turker are employed to verify the validation and test set of our data. Now that the person-box grounding links and candidate person boxes have been checked by workers in VCR, we assume the correctness of them are guaranteed. The two focuses of our verification are ambiguity of grounding links and grammar mistakes or typos in transformed statement. The ambiguity means that in some samples the person mentioned in statement can refer to multiple person boxes in the image. For example, in Fig. \ref{fig:turker} (a), the description is ``PERSONX'' is quite shocked''. However, both ``PERSON0'' and  ``PERSON1''  are quite shocked in the image. So the ground-truth ``PERSON1'' is not complete and this sample is ambiguous. Even though it's acceptable that the noise brought by incomplete links  exists in the training,  we hope to remove it from validation and test set to make sure a precise evaluation. By annotation of workers, only $29\%$ data are ambiguous to certain degree. For grammar mistakes or typos, it's because questions and answers in VCR are in free-form, and finite rules may not cover all cases. As shown in  Fig. \ref{fig:turker} (b), ``PERSONX are ' s arms raised'' is wrong due to the lack of consideration of genitive cases in rule design. Through annotation, we find only $3\%$ data having grammar mistakes or typos.

We remove the ambiguous and unreadable samples. Then, the validation/test set shrinks to 4.9k/4.9k samples, while the training set remains 120k. We pay the workers $0.05\$$ per sample to make their wage $12\$$ to $15\$$ per hour. In order to obtain high quality data, only workers that have finished more than 400 HITs with a decent approval rate of 96\% are allowed for our annotation, which gives us around 90\% agreement in identifying the most likely referred person.

\subsection{Data Analysis}

\label{sec:appendix}
\begin{figure}[t]
        \includegraphics[width=0.31\textheight]{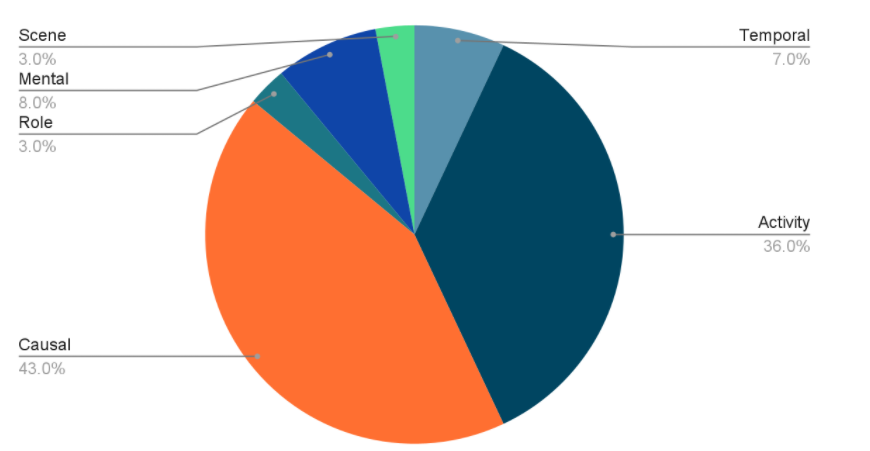}
        \caption{Commonsense Type Analysis}
\label{fig:commonsense_type}
\end{figure}

\paragraph{Commonsense Types}
Our dataset covers plentiful daily scenarios with an enormous diversity in commonsense types. We classify the commonsense types according to the templates. As shown in Fig. \ref{fig:commonsense_type}, $43\%$ samples involve causal commonsense,  $36\%$ samples are related to highly semantic activity commonsense. For some categories, such as causal and temporal, we can further find sub-categories. Causal commonsense can be either causal inference or causal explanation. Temporal commonsense includes before, after and conditional commonsense. Some examples are shown in Fig. \ref{fig:motivation} and more can be found in experiment section.

\paragraph{Comparison with Other Grounding Datasets}
Tab. \ref{tab:datasets} exhibits a comparison of our dataset to previous visual grounding datasets. \textit{HumanCog} is the only one specializing on human-centric commonsense grounding. \textit{Who's Waldo} \cite{cui2021s} is most similar to us, in that both focus on grounding persons. Nevertheless, their samples are crawled from news, where the descriptions are mostly about low-level human actions that are visible straight from images, seldom requiring extra hop of inference and commonsense knowledge.

\section{Method}
\begin{figure}[t]
        \includegraphics[width=0.32\textheight]{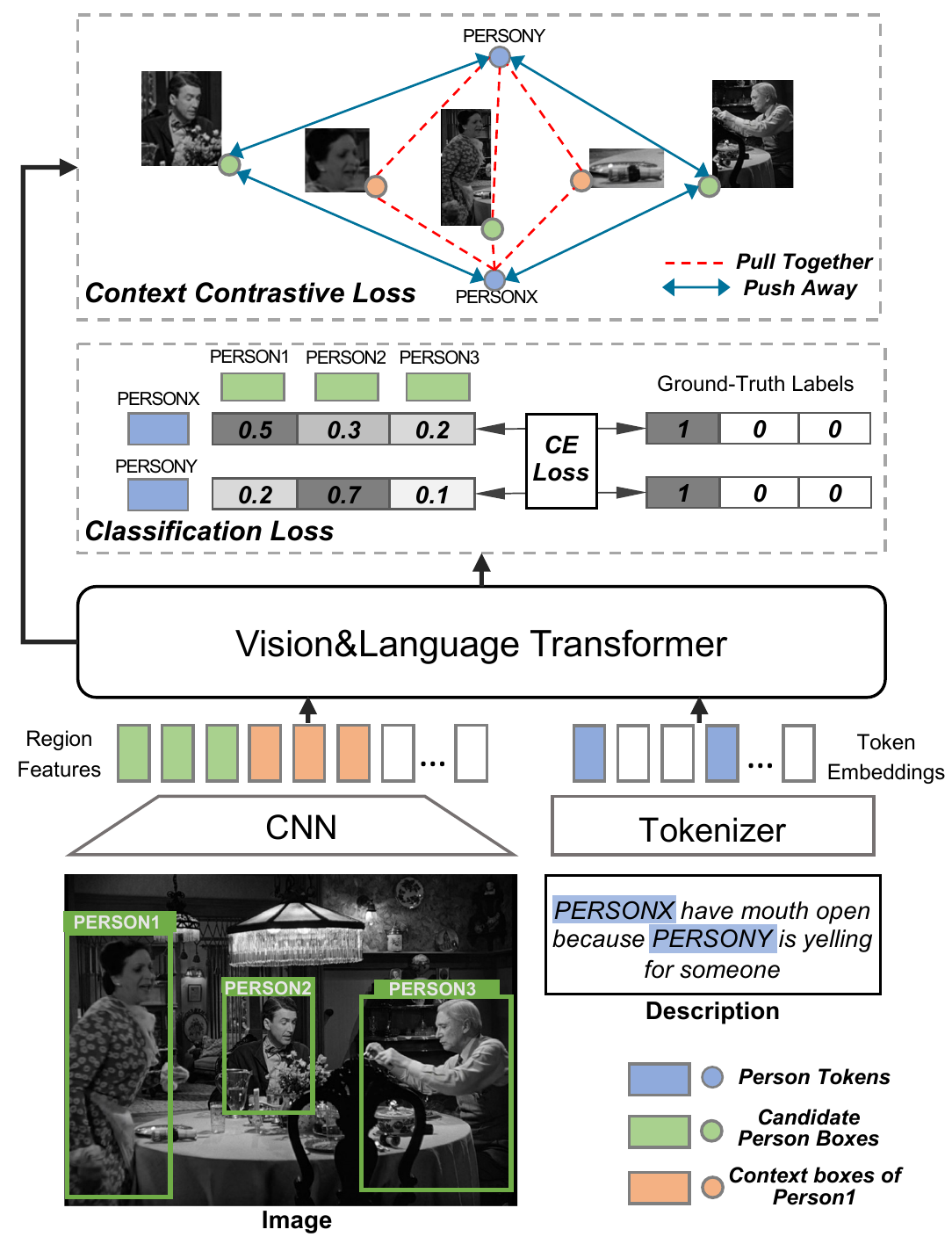}
        \caption{Diagram of our context-object-aware method}
\label{fig:method}
\end{figure}

In this section, we introduce a context-object-aware method as a strong baseline to solve the task. To jointly encode vision and text input, our architecture is built on a pre-trained vision-language Transformer, UNITER \cite{chen2019uniter} for its generality, which will be covered in Sec. \ref{sec:method_1}. In Sec. \ref{sec:method_2}, we introduce the classification loss and the proposed context contrastive loss that can facilitate the visual commonsense learning between the human-object interaction. The diagram of method is shown in Fig. \ref{fig:method}

\subsection{Architecture}
\label{sec:method_1}
\paragraph{Visual Input} Given the image $I$ and candidate person boxes $\boldsymbol{r}$, we follow UNITER \cite{chen2019uniter} to use the Faster R-CNN \cite{ren2015faster} to extract the pooled ROI features of each region $r_i$ as visual features. Location features are encoded by a 7-dimensional vector, $[x_1, y_1, x_2, y_2, w, h, w*h]$\footnote{[normalized top/left/bottom/right coordinates, width, height, area.]}. Visual and location features are transformed into the same dimension through two FC layers, and are then summed up and normalized by a LN, as the input features of each region. Additionally, to complement person representation and enrich the visual scene understanding, we take extra detected object proposals\footnote{Objectness Threshold is 0.2, Max. No. of object is 100.} $\boldsymbol{r}^{'}$ by Faster R-CNN and append their features together with candidate person boxes as input. It's validated in experiments that the additional proposals are essentially helpful to this task.
\paragraph{Textual Input} We tokenize the input description into WordPieces \cite{wu2016google}. The word embeddings and position embeddings are summed up and normalized by a LN, as input text features. As for person links in description, following VL-BERT \cite{su2019vl}, we replace them with random neutral names, \textit{e.g.,} James or Mary.  Compared with initializing new embeddings, it can better utilize the pre-training knowledge.
\paragraph{Transformer Encoder} The visual and textual features are input into the Transformer \cite{chen2019uniter}, pre-trained with 9.5M image-caption pairs. The self-attention layers inside enable the contextualization of the two modalities. We take the hidden layers' features for loss calculation.

\subsection{Loss Function}
\label{sec:method_2}
\paragraph{Classification Loss} 
We treat the task as a classification problem, where each person link $t_i$  in description  $\boldsymbol{t}$ should be classified into the ground-truth person box $r_j$ out of N candidate person boxes $\boldsymbol{r}$. In that way, we transform the features of candidate person boxes into classifier weights and apply a cross-entropy loss $\mathcal{L}_{cls}$:
\begin{align*}
& Q(i,j) = f(t_i)W_1\times(f(r_j)W_2)^T \\
& \mathcal{L}_{cls} = - \frac{1}{k}\sum_{i=1}^{k}\mathrm{log}({Softmax(Q(i,:))}),
\end{align*}
where $f(\cdot)$ denotes the final layer's feature output, $W_1$ and $W_2$ are linear weight matrices, and $k$ is the number of person links in description.
\paragraph{Context Contrastive Loss}
Although the classification loss is straightforward to model human-human interaction, the relationship between detected object proposals and persons is not fully exploited, \textit{i.e.}, the human-object interaction. The surrounding objects can provide plentiful and distinctive semantics to the persons, which is essential to diversify persons and identify ground-truth (GT) persons from other persons. In response to that, we propose a context contrastive loss, where context objects pertaining to GT persons are regarded as positive instances and their features are aligned with corresponding person embeddings in text. More specifically, we pull the person links in description closer to context objects pertaining to corresponding GT persons and push them away from other negative persons in feature space. At first, for person link $t_i$ in text, whose GT person box is $r_j$ in image, we define the pertaining context objects $C(i)$ as those detected boxes that have higher IoU scores with GT person and lower IoU scores with other persons: 

\begin{equation*}
\begin{split}
C(i) & = \{r_{c}^{'}| r_{c}^{'} \in \boldsymbol{r}^{'} \mbox{~and~} IoU(r_{c}^{'}, r_j)>T_1 \\
\mbox{~and~} & max(IoU(r_{c}^{'}, \boldsymbol{r}\backslash r_j))<T_2, \{t_i, r_j\} \in \mathcal{L}\},
\end{split}
\end{equation*}

\noindent where $T_1$ and $T_2$ are two thresholds as hyper-parameters. Then we further include GT person box $r_j$ also in the positive instances, $P(i) = C(i) \vee r_j$. The negative instances are other person boxes $N(i)= \boldsymbol{r}\backslash r_j$ . Contrastive loss has been widely studied in recent works, \textit{e.g.}, InfoNCE \cite{oord2018representation} and its related extensions \cite{he2020momentum, khosla2020supervised, radford2021learning, jia2021scaling}. To realize contrastive learning in our scenario, considering the pertaining objects that have more overlap with GT person $r_j$ tend to be semantically more aligned, we further utilize their IoU scores as the weights in proposed context contrastive loss:
{\setlength{\mathindent}{0.7cm}
\begin{equation*}
    \noindent \mathcal{L}_{con} = - \frac{1}{k}\sum_{i=1}^{k} \mathcal{L}_{con}^{i} 
\end{equation*}}
\begin{equation*}
\begin{split}
     \mathcal{L}_{con}^{i} & =  
    \sum_{r_{p} \in P(i)} \frac{IoU(r_p, r_j)}{|P(i)|} \cdot \\
    & \mathrm{log} \frac{\mathrm{exp}(g_l(t_i)\cdot g_l(r_{p}) / \tau)}{\sum_{r_{n} \in \mathcal{N}(i) \vee P(i)} \mathrm{exp}(g_l(t_i) \cdot g_l(r_{n}) / \tau)},
\end{split}
\end{equation*}
where $\tau$ is the temperature and $g_l(\cdot)$ denotes the encoded feature of $l$-th hidden layer from the last ($l$=3 in our case). Through context contrastive loss, we encourage the person link representations more similar to the correct persons and contextual neighbors of them, and more distinguished from other persons.  

In training, two losses are ensembled together with a coefficient $\lambda$ to adjust the significance of contrastive loss, which is shown as follows. In experiments, we find $\lambda=1$ already gives satisfactory performance. In inference, we take the classification logits and select the box with the highest score as prediction, regardless of the contrastive part.  
\begin{equation*}
    \mathcal{L}_{train} = \mathcal{L}_{cls} + \lambda \mathcal{L}_{con}.
\end{equation*}

\section{Results}
For evaluation, we comprehensively study the effect of various methods on this task. We first introduce previous state-of-the-art (SOTA) methods and compare them with the proposed context-object-aware model (Sec. \ref{sec:exp_sota}). And we conduct ablation experiments to quantify the importance of different components in this task (Sec. \ref{sec:exp_ablation}). Further, we provide qualitative results for analysis (Sec. \ref{sec:exp_qual}). 

\subsection{Experimental Setup}
We compute accuracy of predicted person boxes for all mentioned persons in descriptions as the evaluation metric. For the visual features, if not specified, we use an off-the-shelf pre-trained Faster-RCNN \cite{anderson2018bottom} to extract the region features of both person candidate boxes and detected boxes from image.   As for the training of proposed context-object-aware model, pre-trained weight of UNITER is loaded as the initialization. AdamW  \cite{loshchilov2017decoupled} optimizer is used with a learning rate of 6e-5. Following UNITER, we use dynamic sequence length to batch the samples by their number of input tokens, so that padding is reduced and training is speeded up. We set the batch size to  4000 and train 4k steps. Our model is implemented in Pytorch and trained with 4 TITAN RTX GPUs. \cite{paszke2019pytorch}

\begin{table}[t]
\begin{center}
\renewcommand\tabcolsep{3pt}
\scalebox{0.82}{
\begin{tabular}{cccccc}
    \toprule
   \multirow{2}{*}{Model} &\multirow{2}{*}{Model-size}  &  \# Pre-training  & \multirow{2}{*}{Acc.}\\
    &  & Pairs & \\
    \midrule
    Random & - & - & 30.9 \\ 
    \midrule
    B$\rightarrow$ S & - & - & 39.2 \\ 
    L$\rightarrow$ R & - & - & 31.0 \\
    L$\rightarrow$ R (Biggest) & - & - & 39.6 \\
    \midrule
    BAN(LSTM) & - & - & 56.2 $\pm 0.36$ \\
    BAN(BERT) & - & - & 62.8 $\pm 0.32$ \\
    \midrule
    \multirow{2}{*}{VL-BERT} & base & 3.3M & 67.4 $\pm 0.27$ \\
     & large & 3.3M & 68.2 $\pm 0.22$ \\
    \multirow{2}{*}{VILLA} & base & 9.5M & 68.1 $\pm 0.56$ \\
     & large & 9.5M & 68.5 $\pm 0.52$ \\
    \multirow{2}{*}{UNITER} & base & 9.5M & 67.9 $\pm 0.29$ \\
     & large & 9.5M & \underline{68.9} $\pm 0.31$ \\
     \midrule
     Ours & large & 9.5M & \textbf{69.8} $\pm 0.23$ \\
     \midrule
     Human & - & - & 92.3 \\
    \bottomrule
\end{tabular}
}
\end{center}
\caption{Comparison against previous methods}
\label{tab:sota}
\end{table}

\begin{table}[t]
\begin{center}
\renewcommand\tabcolsep{3pt}
\scalebox{0.95}{
\begin{tabular}{lc}
    \toprule
   Model&Acc.(\%)\\
    \midrule
    Ours & 69.8 \\ 
    \midrule
    $-$ Context Contrastive Loss & 68.9\\
    $-$ Detected Context Objects & 66.9 \\
    $-$ Pre-training on Image-Text Pairs & 64.5 \\
    \bottomrule
\end{tabular}
}
\end{center}
\caption{Ablation of different components in our method}
\label{tab:ablation}
\end{table}

\subsection{Compared with Previous Methods}
\label{sec:exp_sota}
Several previous visual grounding methods are implemented to be compared with our context-object-aware model, including heuristic methods,  models w/o image-text pre-training and models w/ image-text pre-training. 

\noindent \textbf{Heuristic Methods}  Similar to \cite{cui2021s}, we probe the biases in dataset by several hand-crafted heuristics. To be more specific, we assign the persons mentioned in description from left to right to person candidate boxes in image that are sorted based on following heuristics: (1) big to small with decreasing areas; (2) left to right according to the upper-left coordinates (3) left to right with only top-$k$ biggest boxes, where $k$ is the number of mentioned persons in description.

\noindent \textbf{Human Evaluation}  We go through the test set and obtained a 92.3\% accuracy with human evaluation. It can be treated as an reasonable upperbound of current machine models. 

\noindent \textbf{Methods w/o Image-Text Pre-training} In previous works of non-pretrained vision\&Language models, we choose BAN \cite{kim2018bilinear} for its superior performance on visual grounding and other downstream tasks. BAN extracts visual features and text features, and then fuse two modalities by a bilinear attention network. A classification loss as in Sec. \ref{sec:method_2} is applied afterward to do the classification. In our implementation, text feature can be extracted either by a LSTM \cite{hochreiter1997long} module, as in the original paper, or a pre-trained BERT \cite{devlin2018bert} module. We name those two BAN(LSTM) and BAN(BERT).

\noindent\textbf{Methods w/ Image-Text Pre-training} Recently, there have been a lot research interests in vision\&language pre-training models for their effectiveness and generalizability. We implement widely-used VL-BERT~\cite{su2019vl}, UNITER~\cite{chen2019uniter} and VILLA~\cite{gan2020large}, whose numbers of pre-training image-caption pairs range from 3.3M to 9.5M.  Similarly, we also apply the classification loss for training.

The full experimental result, including random guessing, is shown in Tab. \ref{tab:sota}. By comparing the heuristic methods with random guessing, it's found that the strongest heuristic can only improve 9\%, which indicates the spatial bias and area bias is not severe in our dataset. From the comparison of models w/ and w/o image-text pre-training, we find that, in general, larger pre-training data can bring higher performance. Last but not least, proposed context-object-aware method can further outperform UNITER by 0.9\%, reaching a final performance of 69.8\% and establishing a strong baseline.


\begin{figure*}[t]
        \centering
        \includegraphics[width=0.6\textheight]{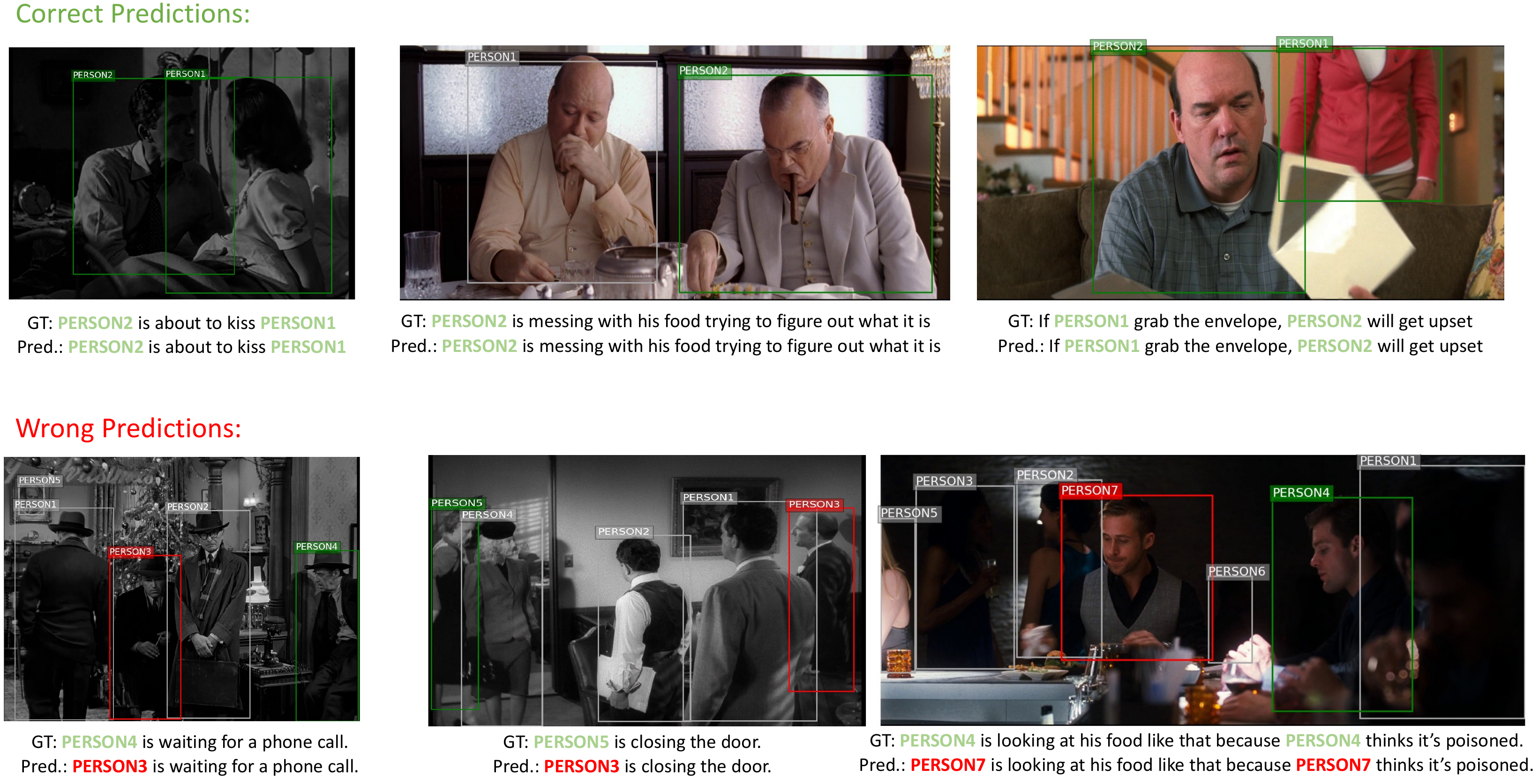}
        \caption{Qualitative Results. GT denotes ground-truth links in description. Pred. denotes predicted links.}
\label{fig:qual}
\end{figure*}

\subsection{Ablation Study}
\label{sec:exp_ablation}
\paragraph{Components}
We further validate the effectiveness of different components in our method. In Tab. \ref{tab:ablation}, we present the results. If the multi-modal Transformer weight is initialized from a BERT model without image-caption pre-training, the performance drops by 5.3\%, which is in line with the finding in previous section that image-text pre-training is beneficial. Removing the context contrastive loss and the detected context objects also  bring a performance degrade of 0.9\% and 2.9\% respectively. It highlights the importance of incorporating contextual objects and enhancing human-object interactions. In summary, both strong image-text fusion and effective  human-object visual commonsense modeling are crucial for this task, which suggests the avenues to future works. 
\paragraph{Hyperparameters} 
We ablated where to insert the context contrastive loss. We found taking the output of 3rd Transformer layer from the last to compute context contrastive loss performs the best. Others bring at most 0.6\% performance drop. We infer that contrastive and classification loss are not fully complementary when added to the same layer. Putting the auxiliary loss earlier helps the last several layers specialize in the classification target.

\subsection{Qualitative Results}
\label{sec:exp_qual}
In Fig.\ref{fig:qual}, we present qualitative results of our method. Ours can correctly ground to the GT person boxes in many cases. In the second correct samples, our model can predict it's PERSON2 instead of PERSON1 probably by the facial expression and the food in front of him. Our model can also work well under the complex causal scenarios, such as the third correct sample. Sometimes, we fail on those samples where the visual information might be incomplete, blurry or misleading. For example, in the first wrong sample, the object PERSON3 is holding a cell phone from appearance, which might mislead the model. And in the 2nd wrong sample, the door is unseen from the image. Some too complex descriptions that require detailed and subtle observation may also cause failure. In the 3rd wrong sample, the PERSON7 is actually hiding something while the PERSON4 is stirring his food, making models hard to tell.
\section{Conclusion}
In this work, we present a new task: \emph{human-centric commonsense grounding}, where machines are required to ground the persons mentioned in a commonsensical description. Correspondingly, we collect a new dataset for training and evaluation. Also, we proposed a context-object-aware method that exploits background context objects via context contrastive loss for a strong vision-language understanding. Through detailed analysis, we find there is still an ample room for improvement and we point out the potential directions for further works. 

\section*{Acknowledgement}
This work is supported by DARPA MCS program under Cooperative Agreement N66001-19-2-4032.

\section*{Limitations}
One limitation of our work is that the training data might be noisy compared with validation and test data. But according to the statistics summarized from validation and test data, we only have around 3\% unreadable samples and 29\% ambiguous samples in the training set, which is acceptable. It's also a design choice to include ambiguous samples as we would like to test if a model can generalize well by learning from noisy data. The noisy setting simulates the process that humans learn from ambiguous examples rather than heavily curated data. And our experimental results prove that trained with noisy data, machines can still greatly outperform random guess.  
\bibliography{anthology,custom}

\begin{thebibliography}{40}
\expandafter\ifx\csname natexlab\endcsname\relax\def\natexlab#1{#1}\fi

\bibitem[{Anderson et~al.(2018)Anderson, He, Buehler, Teney, Johnson, Gould,
  and Zhang}]{anderson2018bottom}
Peter Anderson, Xiaodong He, Chris Buehler, Damien Teney, Mark Johnson, Stephen
  Gould, and Lei Zhang. 2018.
\newblock Bottom-up and top-down attention for image captioning and visual
  question answering.
\newblock In \emph{Proceedings of the IEEE conference on computer vision and
  pattern recognition}, pages 6077--6086.

\bibitem[{Chen et~al.(2019)Chen, Li, Yu, El~Kholy, Ahmed, Gan, Cheng, and
  Liu}]{chen2019uniter}
Yen-Chun Chen, Linjie Li, Licheng Yu, Ahmed El~Kholy, Faisal Ahmed, Zhe Gan,
  Yu~Cheng, and Jingjing Liu. 2019.
\newblock Uniter: Learning universal image-text representations.

\bibitem[{Cui et~al.(2021)Cui, Khandelwal, Artzi, Snavely, and
  Averbuch-Elor}]{cui2021s}
Yuqing Cui, Apoorv Khandelwal, Yoav Artzi, Noah Snavely, and Hadar
  Averbuch-Elor. 2021.
\newblock Who's waldo? linking people across text and images.
\newblock In \emph{Proceedings of the IEEE/CVF International Conference on
  Computer Vision}, pages 1374--1384.

\bibitem[{Demszky et~al.(2018)Demszky, Guu, and
  Liang}]{demszky2018transforming}
Dorottya Demszky, Kelvin Guu, and Percy Liang. 2018.
\newblock Transforming question answering datasets into natural language
  inference datasets.
\newblock \emph{arXiv preprint arXiv:1809.02922}.

\bibitem[{Deng et~al.(2021)Deng, Yang, Chen, Zhou, and Li}]{deng2021transvg}
Jiajun Deng, Zhengyuan Yang, Tianlang Chen, Wengang Zhou, and Houqiang Li.
  2021.
\newblock Transvg: End-to-end visual grounding with transformers.
\newblock In \emph{Proceedings of the IEEE/CVF International Conference on
  Computer Vision}, pages 1769--1779.

\bibitem[{Devlin et~al.(2018)Devlin, Chang, Lee, and
  Toutanova}]{devlin2018bert}
Jacob Devlin, Ming-Wei Chang, Kenton Lee, and Kristina Toutanova. 2018.
\newblock Bert: Pre-training of deep bidirectional transformers for language
  understanding.
\newblock \emph{arXiv preprint arXiv:1810.04805}.

\bibitem[{Fukui et~al.(2016)Fukui, Park, Yang, Rohrbach, Darrell, and
  Rohrbach}]{fukui2016multimodal}
Akira Fukui, Dong~Huk Park, Daylen Yang, Anna Rohrbach, Trevor Darrell, and
  Marcus Rohrbach. 2016.
\newblock Multimodal compact bilinear pooling for visual question answering and
  visual grounding.
\newblock \emph{arXiv preprint arXiv:1606.01847}.

\bibitem[{Gan et~al.(2020)Gan, Chen, Li, Zhu, Cheng, and Liu}]{gan2020large}
Zhe Gan, Yen-Chun Chen, Linjie Li, Chen Zhu, Yu~Cheng, and Jingjing Liu. 2020.
\newblock Large-scale adversarial training for vision-and-language
  representation learning.
\newblock \emph{Advances in Neural Information Processing Systems},
  33:6616--6628.

\bibitem[{He et~al.(2020)He, Fan, Wu, Xie, and Girshick}]{he2020momentum}
Kaiming He, Haoqi Fan, Yuxin Wu, Saining Xie, and Ross Girshick. 2020.
\newblock Momentum contrast for unsupervised visual representation learning.
\newblock In \emph{Proceedings of the IEEE/CVF conference on computer vision
  and pattern recognition}, pages 9729--9738.

\bibitem[{Hochreiter and Schmidhuber(1997)}]{hochreiter1997long}
Sepp Hochreiter and J{\"u}rgen Schmidhuber. 1997.
\newblock Long short-term memory.
\newblock \emph{Neural computation}, 9(8):1735--1780.

\bibitem[{Hu et~al.(2017)Hu, Rohrbach, Andreas, Darrell, and
  Saenko}]{hu2017modeling}
Ronghang Hu, Marcus Rohrbach, Jacob Andreas, Trevor Darrell, and Kate Saenko.
  2017.
\newblock Modeling relationships in referential expressions with compositional
  modular networks.
\newblock In \emph{Proceedings of the IEEE conference on computer vision and
  pattern recognition}, pages 1115--1124.

\bibitem[{Jia et~al.(2021)Jia, Yang, Xia, Chen, Parekh, Pham, Le, Sung, Li, and
  Duerig}]{jia2021scaling}
Chao Jia, Yinfei Yang, Ye~Xia, Yi-Ting Chen, Zarana Parekh, Hieu Pham, Quoc Le,
  Yun-Hsuan Sung, Zhen Li, and Tom Duerig. 2021.
\newblock Scaling up visual and vision-language representation learning with
  noisy text supervision.
\newblock In \emph{International Conference on Machine Learning}, pages
  4904--4916. PMLR.

\bibitem[{Kamath et~al.(2021)Kamath, Singh, LeCun, Synnaeve, Misra, and
  Carion}]{kamath2021mdetr}
Aishwarya Kamath, Mannat Singh, Yann LeCun, Gabriel Synnaeve, Ishan Misra, and
  Nicolas Carion. 2021.
\newblock Mdetr-modulated detection for end-to-end multi-modal understanding.
\newblock In \emph{Proceedings of the IEEE/CVF International Conference on
  Computer Vision}, pages 1780--1790.

\bibitem[{Kazemzadeh et~al.(2014)Kazemzadeh, Ordonez, Matten, and
  Berg}]{KazemzadehOrdonezMattenBergEMNLP14}
Sahar Kazemzadeh, Vicente Ordonez, Mark Matten, and Tamara~L. Berg. 2014.
\newblock Referit game: Referring to objects in photographs of natural scenes.
\newblock In \emph{EMNLP}.

\bibitem[{Khosla et~al.(2020)Khosla, Teterwak, Wang, Sarna, Tian, Isola,
  Maschinot, Liu, and Krishnan}]{khosla2020supervised}
Prannay Khosla, Piotr Teterwak, Chen Wang, Aaron Sarna, Yonglong Tian, Phillip
  Isola, Aaron Maschinot, Ce~Liu, and Dilip Krishnan. 2020.
\newblock Supervised contrastive learning.
\newblock \emph{Advances in Neural Information Processing Systems},
  33:18661--18673.

\bibitem[{Kim et~al.(2018)Kim, Jun, and Zhang}]{kim2018bilinear}
Jin-Hwa Kim, Jaehyun Jun, and Byoung-Tak Zhang. 2018.
\newblock Bilinear attention networks.
\newblock \emph{Advances in Neural Information Processing Systems}, 31.

\bibitem[{Lei et~al.(2020)Lei, Yu, Berg, and Bansal}]{lei2020more}
Jie Lei, Licheng Yu, Tamara~L Berg, and Mohit Bansal. 2020.
\newblock What is more likely to happen next? video-and-language future event
  prediction.
\newblock \emph{arXiv preprint arXiv:2010.07999}.

\bibitem[{Li et~al.(2019)Li, Yatskar, Yin, Hsieh, and Chang}]{li2019visualbert}
Liunian~Harold Li, Mark Yatskar, Da~Yin, Cho-Jui Hsieh, and Kai-Wei Chang.
  2019.
\newblock Visualbert: A simple and performant baseline for vision and language.
\newblock \emph{arXiv preprint arXiv:1908.03557}.

\bibitem[{Liao et~al.(2020)Liao, Liu, Li, Wang, Chen, Qian, and
  Li}]{liao2020real}
Yue Liao, Si~Liu, Guanbin Li, Fei Wang, Yanjie Chen, Chen Qian, and Bo~Li.
  2020.
\newblock A real-time cross-modality correlation filtering method for referring
  expression comprehension.
\newblock In \emph{Proceedings of the IEEE/CVF Conference on Computer Vision
  and Pattern Recognition}, pages 10880--10889.

\bibitem[{Lin et~al.(2014)Lin, Maire, Belongie, Hays, Perona, Ramanan,
  Doll{\'a}r, and Zitnick}]{lin2014microsoft}
Tsung-Yi Lin, Michael Maire, Serge Belongie, James Hays, Pietro Perona, Deva
  Ramanan, Piotr Doll{\'a}r, and C~Lawrence Zitnick. 2014.
\newblock Microsoft coco: Common objects in context.
\newblock In \emph{European conference on computer vision}, pages 740--755.
  Springer.

\bibitem[{Liu et~al.(2019)Liu, Liu, Bai, and Yuille}]{liu2019clevr}
Runtao Liu, Chenxi Liu, Yutong Bai, and Alan Yuille. 2019.
\newblock Clevr-ref+: Diagnosing visual reasoning with referring expressions.
\newblock \emph{arXiv preprint arXiv:1901.00850}.

\bibitem[{Liu et~al.(2020)Liu, Wan, Zhu, and He}]{liu2020learning}
Yongfei Liu, Bo~Wan, Xiaodan Zhu, and Xuming He. 2020.
\newblock Learning cross-modal context graph for visual grounding.
\newblock In \emph{Proceedings of the AAAI Conference on Artificial
  Intelligence}, volume~34, pages 11645--11652.

\bibitem[{Loshchilov and Hutter(2017)}]{loshchilov2017decoupled}
Ilya Loshchilov and Frank Hutter. 2017.
\newblock Decoupled weight decay regularization.
\newblock \emph{arXiv preprint arXiv:1711.05101}.

\bibitem[{Lu et~al.(2019)Lu, Batra, Parikh, and Lee}]{lu2019vilbert}
Jiasen Lu, Dhruv Batra, Devi Parikh, and Stefan Lee. 2019.
\newblock Vilbert: Pretraining task-agnostic visiolinguistic representations
  for vision-and-language tasks.
\newblock \emph{Advances in neural information processing systems}, 32.

\bibitem[{Luo and Shakhnarovich(2017)}]{luo2017comprehension}
Ruotian Luo and Gregory Shakhnarovich. 2017.
\newblock Comprehension-guided referring expressions.
\newblock In \emph{Proceedings of the IEEE Conference on Computer Vision and
  Pattern Recognition}, pages 7102--7111.

\bibitem[{Mao et~al.(2016)Mao, Huang, Toshev, Camburu, Yuille, and
  Murphy}]{mao2016generation}
Junhua Mao, Jonathan Huang, Alexander Toshev, Oana Camburu, Alan~L Yuille, and
  Kevin Murphy. 2016.
\newblock Generation and comprehension of unambiguous object descriptions.
\newblock In \emph{Proceedings of the IEEE conference on computer vision and
  pattern recognition}, pages 11--20.

\bibitem[{Oord et~al.(2018)Oord, Li, and Vinyals}]{oord2018representation}
Aaron van~den Oord, Yazhe Li, and Oriol Vinyals. 2018.
\newblock Representation learning with contrastive predictive coding.
\newblock \emph{arXiv preprint arXiv:1807.03748}.

\bibitem[{Park et~al.(2020)Park, Bhagavatula, Mottaghi, Farhadi, and
  Choi}]{park2020visualcomet}
Jae~Sung Park, Chandra Bhagavatula, Roozbeh Mottaghi, Ali Farhadi, and Yejin
  Choi. 2020.
\newblock Visualcomet: Reasoning about the dynamic context of a still image.
\newblock In \emph{In Proceedings of the European Conference on Computer Vision
  (ECCV)}.

\bibitem[{Paszke et~al.(2019)Paszke, Gross, Massa, Lerer, Bradbury, Chanan,
  Killeen, Lin, Gimelshein, Antiga et~al.}]{paszke2019pytorch}
Adam Paszke, Sam Gross, Francisco Massa, Adam Lerer, James Bradbury, Gregory
  Chanan, Trevor Killeen, Zeming Lin, Natalia Gimelshein, Luca Antiga, et~al.
  2019.
\newblock Pytorch: An imperative style, high-performance deep learning library.
\newblock \emph{Advances in neural information processing systems}, 32.

\bibitem[{Plummer et~al.(2015)Plummer, Wang, Cervantes, Caicedo, Hockenmaier,
  and Lazebnik}]{plummer2015flickr30k}
Bryan~A Plummer, Liwei Wang, Chris~M Cervantes, Juan~C Caicedo, Julia
  Hockenmaier, and Svetlana Lazebnik. 2015.
\newblock Flickr30k entities: Collecting region-to-phrase correspondences for
  richer image-to-sentence models.
\newblock In \emph{Proceedings of the IEEE international conference on computer
  vision}, pages 2641--2649.

\bibitem[{Qiao et~al.(2020)Qiao, Deng, and Wu}]{qiao2020referring}
Yanyuan Qiao, Chaorui Deng, and Qi~Wu. 2020.
\newblock Referring expression comprehension: A survey of methods and datasets.
\newblock \emph{IEEE Transactions on Multimedia}, 23:4426--4440.

\bibitem[{Radford et~al.(2021)Radford, Kim, Hallacy, Ramesh, Goh, Agarwal,
  Sastry, Askell, Mishkin, Clark et~al.}]{radford2021learning}
Alec Radford, Jong~Wook Kim, Chris Hallacy, Aditya Ramesh, Gabriel Goh,
  Sandhini Agarwal, Girish Sastry, Amanda Askell, Pamela Mishkin, Jack Clark,
  et~al. 2021.
\newblock Learning transferable visual models from natural language
  supervision.
\newblock In \emph{International Conference on Machine Learning}, pages
  8748--8763. PMLR.

\bibitem[{Ren et~al.(2015)Ren, He, Girshick, and Sun}]{ren2015faster}
Shaoqing Ren, Kaiming He, Ross Girshick, and Jian Sun. 2015.
\newblock Faster r-cnn: Towards real-time object detection with region proposal
  networks.
\newblock \emph{Advances in neural information processing systems}, 28.

\bibitem[{Su et~al.(2019)Su, Zhu, Cao, Li, Lu, Wei, and Dai}]{su2019vl}
Weijie Su, Xizhou Zhu, Yue Cao, Bin Li, Lewei Lu, Furu Wei, and Jifeng Dai.
  2019.
\newblock Vl-bert: Pre-training of generic visual-linguistic representations.
\newblock \emph{arXiv preprint arXiv:1908.08530}.

\bibitem[{Wang et~al.(2020)Wang, Liu, Li, and Wu}]{wang2020give}
Peng Wang, Dongyang Liu, Hui Li, and Qi~Wu. 2020.
\newblock Give me something to eat: referring expression comprehension with
  commonsense knowledge.
\newblock In \emph{Proceedings of the 28th ACM International Conference on
  Multimedia}, pages 28--36.

\bibitem[{Wu et~al.(2016)Wu, Schuster, Chen, Le, Norouzi, Macherey, Krikun,
  Cao, Gao, Macherey et~al.}]{wu2016google}
Yonghui Wu, Mike Schuster, Zhifeng Chen, Quoc~V Le, Mohammad Norouzi, Wolfgang
  Macherey, Maxim Krikun, Yuan Cao, Qin Gao, Klaus Macherey, et~al. 2016.
\newblock Google's neural machine translation system: Bridging the gap between
  human and machine translation.
\newblock \emph{arXiv preprint arXiv:1609.08144}.

\bibitem[{Yang et~al.(2019)Yang, Li, and Yu}]{yang2019cross}
Sibei Yang, Guanbin Li, and Yizhou Yu. 2019.
\newblock Cross-modal relationship inference for grounding referring
  expressions.
\newblock In \emph{Proceedings of the IEEE/CVF Conference on Computer Vision
  and Pattern Recognition}, pages 4145--4154.

\bibitem[{Yu et~al.(2018)Yu, Lin, Shen, Yang, Lu, Bansal, and
  Berg}]{yu2018mattnet}
Licheng Yu, Zhe Lin, Xiaohui Shen, Jimei Yang, Xin Lu, Mohit Bansal, and
  Tamara~L Berg. 2018.
\newblock Mattnet: Modular attention network for referring expression
  comprehension.
\newblock In \emph{Proceedings of the IEEE Conference on Computer Vision and
  Pattern Recognition}, pages 1307--1315.

\bibitem[{Yu et~al.(2016)Yu, Poirson, Yang, Berg, and Berg}]{yu2016modeling}
Licheng Yu, Patrick Poirson, Shan Yang, Alexander~C Berg, and Tamara~L Berg.
  2016.
\newblock Modeling context in referring expressions.
\newblock In \emph{European Conference on Computer Vision}, pages 69--85.
  Springer.

\bibitem[{Zellers et~al.(2019)Zellers, Bisk, Farhadi, and
  Choi}]{zellers2019recognition}
Rowan Zellers, Yonatan Bisk, Ali Farhadi, and Yejin Choi. 2019.
\newblock From recognition to cognition: Visual commonsense reasoning.
\newblock In \emph{Proceedings of the IEEE/CVF conference on computer vision
  and pattern recognition}, pages 6720--6731.

\end{thebibliography}
\bibliographystyle{acl_natbib}

\appendix


\end{document}